\documentclass{article}
\usepackage{spconf,amsmath,graphicx,booktabs,multirow,float,amssymb,amsfonts,array}
\usepackage{hyperref}

\title{CONFIDENCE-AWARE MULTI-TEACHER KNOWLEDGE DISTILLATION}
\name{Hailin Zhang \qquad Defang Chen \qquad Can Wang$^{\star}$
\thanks{$^{\star}$Corresponding author}
\thanks{This work is supported by National Key R\&D Program of China (Grant No: 2019YFB1600700), the Starry Night Science Fund of Zhejiang University Shanghai Institute for Advanced Study (Grant No: SN-ZJU-SIAS-001) and National Natural Science Foundation of China (Grant No: U1866602).}}
\address{Zhejiang University, China; ZJU-Bangsun Joint Research Center.\\
\{zzzhl, defchern, wcan\}@zju.edu.cn}
\begin{document}
%
\maketitle
\begin{abstract}
{Knowledge distillation is initially introduced to utilize additional supervision from a single teacher model for the student model training. To boost the student performance, some recent variants attempt to exploit diverse knowledge sources from multiple teachers. 
However, existing studies mainly integrate knowledge from diverse sources by averaging over multiple teacher predictions or combining them using other label-free strategies, which may mislead student in the presence of low-quality teacher predictions.
To tackle this problem, we propose Confidence-Aware Multi-teacher Knowledge Distillation (CA-MKD), which adaptively assigns sample-wise reliability for each teacher prediction with the help of ground-truth labels, with those teacher predictions close to one-hot labels assigned large weights.
Besides, CA-MKD incorporates features in intermediate layers to stable the knowledge transfer process. Extensive experiments show our CA-MKD consistently outperforms all compared state-of-the-art methods across various teacher-student architectures. Code is available: \url{https://github.com/Rorozhl/CA-MKD}.
}
\end{abstract}

\section{Introduction}
\label{sec:intro}

Nowadays, deep neural networks have achieved unprecedented success in various applications \cite{he2016deep,silver2017mastering,devlin2019bert}. However, these complex models requiring huge memory footprint and computational resources are difficult to be applied on embedded devices. 
Knowledge distillation (KD) is thus proposed as a model compression technique to resolve this issue, which improves the accuracy of a lightweight student model by distilling the knowledge from a pre-trained cumbersome teacher model \cite{hinton2015distilling}. The transferred knowledge was originally formalized as softmax outputs (soft targets) of the teacher model \cite{hinton2015distilling} and latter extended to the intermediate teacher layers for achieving more promising performance \cite{romero2015fitnet,zagoruyko2017paying,chen2021cross}. 
\begin{figure}[t]
    \centering
    \centerline{\includegraphics[width=0.75\columnwidth]{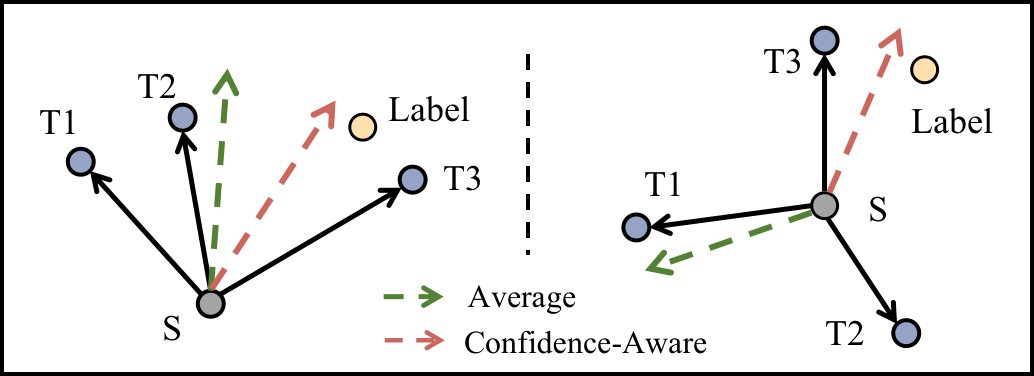}}
    \caption{Comparison of the previous average direction (green line) and our proposed confidence-aware direction (red line).} 
    \label{Figure 1}
    \vspace{-0.45cm}
\end{figure} 

As the wisdom of the masses exceeds that of the wisest individual, some multi-teacher knowledge distillation (MKD) methods are proposed and have been proven to be beneficial \cite{you2017learning, fukuda2017efficient, wu2019multi,du2020agree,kwon2020adaptive}. Basically, they combine predictions from multiple teachers with 
the fixed weight assignment \cite{you2017learning, fukuda2017efficient, wu2019multi} or other various label-free schemes, such as calculating weights based on a optimization problem or entropy criterion \cite{du2020agree, kwon2020adaptive}, etc. 
However, fixed weights fail to differentiate high-quality teachers from low-quality ones \cite{you2017learning, fukuda2017efficient, wu2019multi}, and the other schemes may mislead the student in the presence of low-quality teacher predictions \cite{du2020agree, kwon2020adaptive}. Figure~\ref{Figure 1} provides an intuitive illustration on this issue, where the student trained with the average weighting strategy might deviate from the correct direction once most teacher predictions are biased.  

Fortunately, we actually have ground-truth labels in hand to quantify our confidence about teacher predictions and then filter out low-quality predictions for better student training. To this end, we propose Confidence-Aware Multi-teacher Knowledge Distillation (CA-MKD) to learn sample-wise weights by taking the prediction confidence of teachers into consideration for adaptive knowledge integration. The confidence is obtained based on the cross entropy loss between prediction distributions and ground-truth labels. Compared with previous label-free weighting strategies, our technique enables the student to learn from a relatively correct direction.

Note that our confidence-aware mechanism not only is able to adaptively weight different teacher predictions based on their sample-wise confidence, but also can be extended to the student-teacher feature pairs in intermediate layers. With the help of our generated flexible and effective weights, we could avoid those poor teacher predictions dominating the knowledge transfer process and considerably improve 
the student performance on eight teacher-student architecture combinations (as shown in Table~\ref{table:MKD} and~\ref{table:MKD_d}).

\section{RELATED WORK}
\label{sec:format}

{\bf Knowledge Distillation.} Vanilla KD aims to transfer knowledge from a complex network (teacher) to a simple network (student) with the KL divergence minimization between their softened outputs \cite{ba2014deep,hinton2015distilling}. Mimicking the teacher representations from intermediate layers was latter proposed to explore more knowledge forms \cite{romero2015fitnet, zagoruyko2017paying,ahn2019variational,tian2020contrastive,chen2021cross}. Compared to these methods that require pre-training a teacher, some works simultaneously train multiple students and encourage them to learn from each other instead \cite{lan2018knowledge, chen2020online}. Our technique differs from these online KD methods since we attempt to distill knowledge from multiple pre-trained teachers.
\\
{\bf Multi-teacher Knowledge Distillation.} Rather than employing a single teacher, MKD boosts the effectiveness of distillation by integrating predictions from multiple teachers. 
A bunch of methods are proposed, such as simply assigning average or other fixed weights for different teachers \cite{you2017learning,fukuda2017efficient,wu2019multi}, and calculating the weights based on entropy \cite{kwon2020adaptive}, latent factor \cite{liu2020adaptive} or multi-objective optimization in the gradient space \cite{du2020agree}. However, these label-free strategies may mislead the student training in the presence of low-quality predictions. For instance, entropy-based strategy will prefer models with blind faith since it favors predictions with low variance \cite{kwon2020adaptive}; optimization-based strategy favors majority opinion and will be easily misled by noisy data \cite{du2020agree}. 
In contrast, our CA-MKD quantifies the teacher predictions based on ground-truth labels and further improves the student performance.

\section{METHODOLOGY}

We denote $\mathcal D=\{{\boldsymbol x_{i}}, {\boldsymbol y_{i}} \}^{N}_{i}$ as a labeled training set, $N$ is the number of samples, $K$ is the number of teachers. $F\in \mathbb{R}^{h\times w\times c}$ is the output of the last network block. We denote $\boldsymbol{z}=[z^{1}, ...,z^{C}]$ as the logits output, where $C$ is the category number. The final model prediction is obtained by a softmax function $\sigma\left(z^{c}\right)=\frac{\exp\left(z^{c}/\tau\right)}{\sum_{j}\exp\left(z^{j}/\tau\right)}$ with temperature $\tau$. In the following sections, we will introduce our CA-MKD in detail. 

\subsection{The Loss of Teacher Predictions}
To effectively aggregate the prediction distributions of multiple teachers, we assign different weights which reflects their sample-wise confidence by calculating the cross entropy loss between teacher predictions and ground-truth labels
\begin{equation}
\mathcal L_{CE_{KD}}^{k}=-\sum^{C}_{c=1}y^{c}\log\left(\sigma\left(z_{T_{k}}^{c}\right)\right), 
\end{equation}
\begin{equation}
w^{k}_{KD}=\frac{1}{K-1}\left(1 - \frac{\exp\left(\mathcal L_{CE_{KD}}^{k}\right)}{\sum_{j}\exp\left(\mathcal L_{CE_{KD}}^{j}\right)}\right),
\end{equation}
where $T_{k}$ denotes the $k$th teacher. The less $\mathcal L_{CE_{KD}}^{k}$ corresponds to the larger $w^{k}_{KD}$. The overall teacher predictions are then aggregated with calculated weights
\begin{equation}
\mathcal L_{KD}=-\sum^{K}_{k=1}w^{k}_{KD} \sum^{C}_{c=1} z^{c}_{T_{k}}\log \left(\sigma \left(z^{c}_{S}\right)\right).
\end{equation}

According to the above formulas, the teacher whose prediction is closer to ground-truth labels will be assigned larger weight $w^{k}_{KD}$, since it has enough confidence to make accurate judgement for correct guidance. In contrast, if we simply acquire the weights by calculating the entropy of teacher predictions \cite{kwon2020adaptive}, the weight will become large when the output distribution is sharp regardless of whether the highest probability category is correct. In this case, those biased targets may misguide the student training and further hurt its distillation performance.
\begin{figure}[t]
    \centering
    \centerline{\includegraphics[width=8cm]{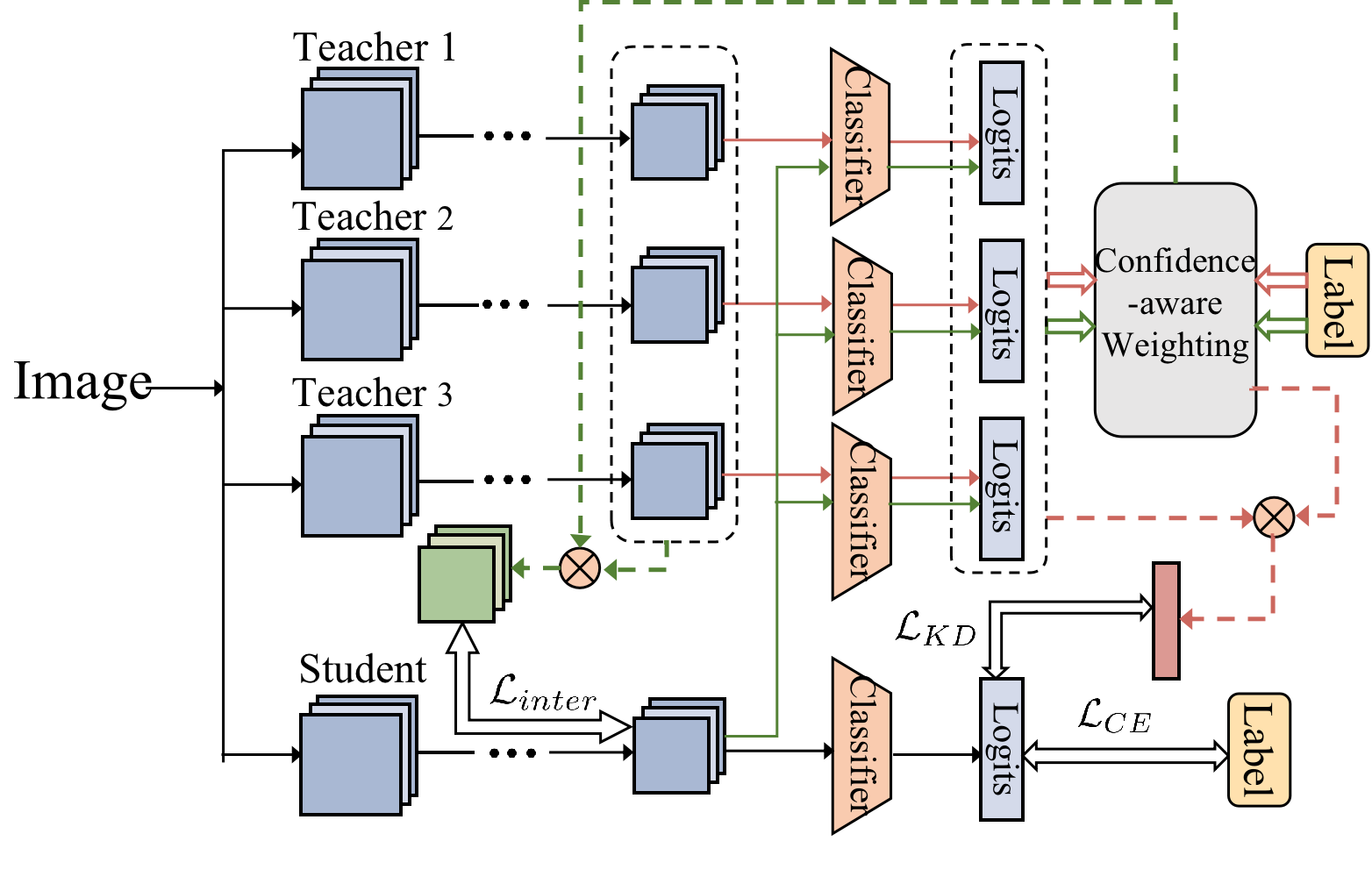}}
    \caption{An overview of our CA-MKD. The weight calculation of 
    teacher predictions and intermediate teacher features are depicted as the red lines and green lines, respectively.} 
    \label{Figure 2}
    \vspace{-0.3cm}
\end{figure}
\begin{table*}[t]
\centering
\caption{Top-1 test accuracy of MKD methods by distilling the knowledge on multiple teachers with the same architectures.}
\label{table:MKD}
\resizebox{0.98\textwidth}{!}{
\begin{tabular}{cccccccc}
\toprule
\multirow{2}*{Teacher}&WRN40-2 &ResNet56 &VGG13 &VGG13 &ResNet32x4 &ResNet32x4 &ResNet32x4 \\
&76.62$\pm$0.26 &73.28$\pm$0.30 &75.17$\pm$0.18 &75.17$\pm$0.18 &79.31$\pm$0.14 &79.31$\pm$0.14 &79.31$\pm$0.14\\
Ensemble &79.62 &76.00 &77.07 &77.07 &81.16 &81.16 &81.16\\
\midrule
\multirow{2}*{Student}&ShuffleNetV1 &MobileNetV2 &VGG8 &MobileNetV2 &ResNet8x4 &ShuffleNetV2 &VGG8 \\
&71.70$\pm$0.43 &65.64$\pm$0.19 &70.74$\pm$0.40 &65.64$\pm$0.19 &72.79$\pm$0.14 &72.94$\pm$0.24 &70.74$\pm$0.40\\
\midrule
AVER \cite{you2017learning} &76.30$\pm$0.25 &70.21$\pm$0.10 &74.07$\pm$0.23 &68.91$\pm$0.35 &74.99$\pm$0.24 &75.87$\pm$0.19 &73.26$\pm$0.39\\
FitNet-MKD \cite{romero2015fitnet} &76.59$\pm$0.17 &70.69$\pm$0.56 &73.97$\pm$0.22 &68.48$\pm$0.07 &74.86$\pm$0.21 &76.09$\pm$0.13 &73.27$\pm$0.19\\
EBKD \cite{kwon2020adaptive} &76.61$\pm$0.14 &70.91$\pm$0.22 &74.10$\pm$0.27 &68.24$\pm$0.82 &75.59$\pm$0.15 &76.41$\pm$0.12 &73.60$\pm$0.22\\
AEKD \cite{du2020agree} &76.34$\pm$0.24 &70.47$\pm$0.15 &73.78$\pm$0.03 &68.39$\pm$0.50 &74.75$\pm$0.28 &75.95$\pm$0.20 &73.11$\pm$0.27\\
CA-MKD &\bf77.94$\pm$0.31 &\bf71.38$\pm$0.02 &\bf74.30$\pm$0.16 &\bf69.41$\pm$0.20 &\bf75.90$\pm$0.13 &\bf77.41$\pm$0.14 &\bf75.26$\pm$0.32\\
\bottomrule
\end{tabular}}
\end{table*}

\begin{table}[t]
\centering
\caption{Top-1 test accuracy of CA-MKD compared to single-teacher knowledge distillation methods.}
\label{table:single}
\resizebox{0.98\columnwidth}{!}{
\begin{tabular}{cccc}
\toprule
\multirow{2}*{Teacher}&WRN40-2 &ResNet32x4 &ResNet56\\
&76.62$\pm$0.26 &79.31$\pm$0.14 &73.28$\pm$0.30\\
\midrule
\multirow{2}*{Student}&ShuffleNetV1 &VGG8 &MobileNetV2\\
&71.70$\pm$0.19 &70.74$\pm$0.40
&65.64$\pm$0.43\\
\midrule
KD \cite{hinton2015distilling} &75.77$\pm$0.14 &72.90$\pm$0.34 &69.96$\pm$0.14\\
FitNet \cite{romero2015fitnet} &76.22$\pm$0.21 &72.55$\pm$0.66 &69.02$\pm$0.28\\
AT \cite{zagoruyko2017paying} &76.44$\pm$0.38 &72.16$\pm$0.12 &69.79$\pm$0.26\\
VID \cite{ahn2019variational} &76.32$\pm$0.08 &73.09$\pm$0.29 &69.45$\pm$0.17\\
CRD \cite{tian2020contrastive} &76.58$\pm$0.23 &73.57$\pm$0.25 &71.15$\pm$0.44\\
\midrule
CA-MKD &\bf77.94$\pm$0.31 &\bf75.26$\pm$0.13 &\bf71.38$\pm$0.02\\
\bottomrule
\end{tabular}}
\end{table}

\subsection{The Loss of Intermediate Teacher Features}
In addition to KD Loss, inspired by FitNets \cite{romero2015fitnet}, we believe that the intermediate layers are also beneficial for learning structural knowledge, and thus extend our method to intermediate layers for mining more information. The calculation of intermediate feature matching is presented as follows
\begin{equation}
z_{S\rightarrow T_{k}}=W_{T_{k}}h_{S}, 
\end{equation}
\begin{equation}
\mathcal L_{CE_{inter}}^{k}=-\sum^{C}_{c=1}y^{c}\log\left(\sigma\left(z_{S\rightarrow T_{k}}^{c}\right)\right),
\end{equation}
\begin{equation}
w^{k}_{inter}=\frac{1}{K-1}\left(1-\frac{\exp\left(\mathcal L_{CE_{inter}}^{k}\right)}{\sum_{j}\exp\left(\mathcal L_{CE_{inter}}^{j}\right)}\right).
\end{equation}
where $W_{T_{k}}$ is the final classifier of the $k$th teacher. $h_{S}\in \mathbb{R}^{c}$ is the last student feature vector, i.e, $h_{S}=\mathrm{AvgPooling}(F_S)$. $\mathcal L_{CE_{inter}}^{k}$ is obtained by passing $h_{S}$ through each teacher classifier. The calculation of $w^{k}_{inter}$ is similar to that of $w^{k}_{KD}$.

To stable the knowledge transfer process, we design the student to be more focused on imitating the teacher with a similar feature space and $w^{k}_{inter}$ indeed serves as such a similarity measure representing the discriminability of a teacher classifier in the student feature space. The ablation study also shows that utilizing $w^{k}_{inter}$ instead of $w^{k}_{KD}$ for the knowledge aggregation in intermediate layers is more effective.

\begin{equation}
\label{eq:inter}
\mathcal L_{inter}=\sum^{K}_{k=1}w^{k}_{inter}||F_{T_{k}}-r\left(F_{S}\right)||^{2}_{2},
\end{equation}
where $r(\cdot)$ is a function for aligning the student and teacher feature dimensions. The $\ell_2$ loss function is used as distance measure of intermediate features. Finally, the overall training loss between feature pairs will be aggregated by $w^{k}_{inter}$. 

In our work, only the output features of the last block are adopted to avoid incurring too much computational cost. 

\subsection{The Overall Loss Function}
In addition to the aforementioned two losses, a regular cross entropy with the ground-truth labels is calculated
\begin{equation}
\mathcal L_{CE}=-\sum^{C}_{c=1}y^{c}\log\left(\sigma(z_{S}^{c})\right).
\end{equation}
The overall loss function of our CA-MKD is summarize as
\begin{equation}
\mathcal L=\mathcal L_{CE}+\alpha \mathcal L_{KD} +\beta \mathcal L_{inter},
\end{equation}
where $\alpha$ and $\beta$ are hyper-parameters to balance the effect of knowledge distillation and standard cross entropy losses. 

\section{EXPERIMENT}
In this section, we conduct extensive experiments on CIFAR-100 dataset \cite{krizhevsky2009learning} to verify the effectiveness of our proposed CA-MKD.
We adopt eight different teacher-student combinations based on popular neural network architectures. All compared multi-teacher knowledge distillation (MKD) methods use three teachers except for special declarations. 

{\bf Compared Methods.} Besides the na{\"i}ve AVER \cite{you2017learning}, we reimplement a single-teacher based method FitNet \cite{romero2015fitnet} on multiple teachers and denote it as FitNet-MKD. FitNet-MKD will leverage extra information coming from averaged intermediate teacher features. We also reimplement an entropy-based MKD method \cite{kwon2020adaptive}, which has achieved remarkable results in acoustic experiments, on our image classification task and we denote it as EBKD. As for AEKD, we adopt its logits-based version with the author provided code \cite{du2020agree}.

{\bf Hyper-parameters.} All neural networks are optimized by stochastic gradient descent with momentum 0.9, weight decay 0.0001. The batch size is set to 64. As the previous works do \cite{tian2020contrastive,chen2021cross}, the initial learning rate is set to 0.1, except MobileNetV2, ShuffleNetV1 and ShuffleNetV2 are set to 0.05. The learning rate is multiplied by 0.1 at 150, 180 and 210 of the total 240 training epochs. For the sake of fairness, the temperature $\tau$ is set to 4 and the $\alpha$ is set to 1 in all methods. Furthermore, we set the $\beta$ of our CA-MKD to 50 throughout the experiments. 
All results are reported in means and standard deviations over 3 runs with different random seeds.

\subsection{Results on the Same Teacher Architectures}
\begin{table*}[t]
\centering
\caption{Top-1 test accuracy of MKD approaches by distilling the knowledge on multiple teachers with different architectures.}
\label{table:MKD_d}
\resizebox{0.98\textwidth}{!}{
\begin{tabular}{c|ccccc|ccc}
\toprule
VGG8 &AVER &FitNet-MKD &EBKD &AEKD &CA-MKD &ResNet8x4 &ResNet20x4 &ResNet32x4\\
70.74$\pm$0.40 &74.55$\pm$0.24 &74.47$\pm$0.21 &74.07$\pm$0.17 &74.69$\pm$0.29 &\bf75.96$\pm$0.05 &72.79 &78.39 &79.31\\
\bottomrule
\end{tabular}}
\end{table*}

Table~\ref{table:MKD} shows the top-1 accuracy comparison on CIFAR-100. We also include the results of teacher ensemble with the majority voting strategy. We can find that CA-MKD surpasses all competitors cross various architectures. Specifically, compared to the second best method (EBKD), CA-MKD outperforms it with 0.81\% average improvement\footnote{\mbox{Average Improvement}=$\frac{1}{n}\sum_{i}^{n}\left(Acc_{\mathrm{CA-MKD}}^{i}-Acc_{\mathrm{EBKD}}^{i}\right)$, where the accuracies of CA-MKD, EBKD in the $i$-th teacher-student combination are denoted as $Acc_{\mathrm{CA-MKD}}^{i}$, $Acc_{\mathrm{EBKD}}^{i}$, respectively.}, and achieves 1.66\% absolute accuracy improvement in the best case. 

To verify the benefits of diverse information brought by multiple teachers, we compare CA-MKD with some excellent single-teacher based methods. The results in Table~\ref{table:single} show the student indeed has the potential to learn knowledge from multiple teachers, and its accuracy is further improved compared with the single-teacher methods to a certain extent.
\begin{figure}[t]
    \centering
    \centerline{\includegraphics[width=8.5cm]{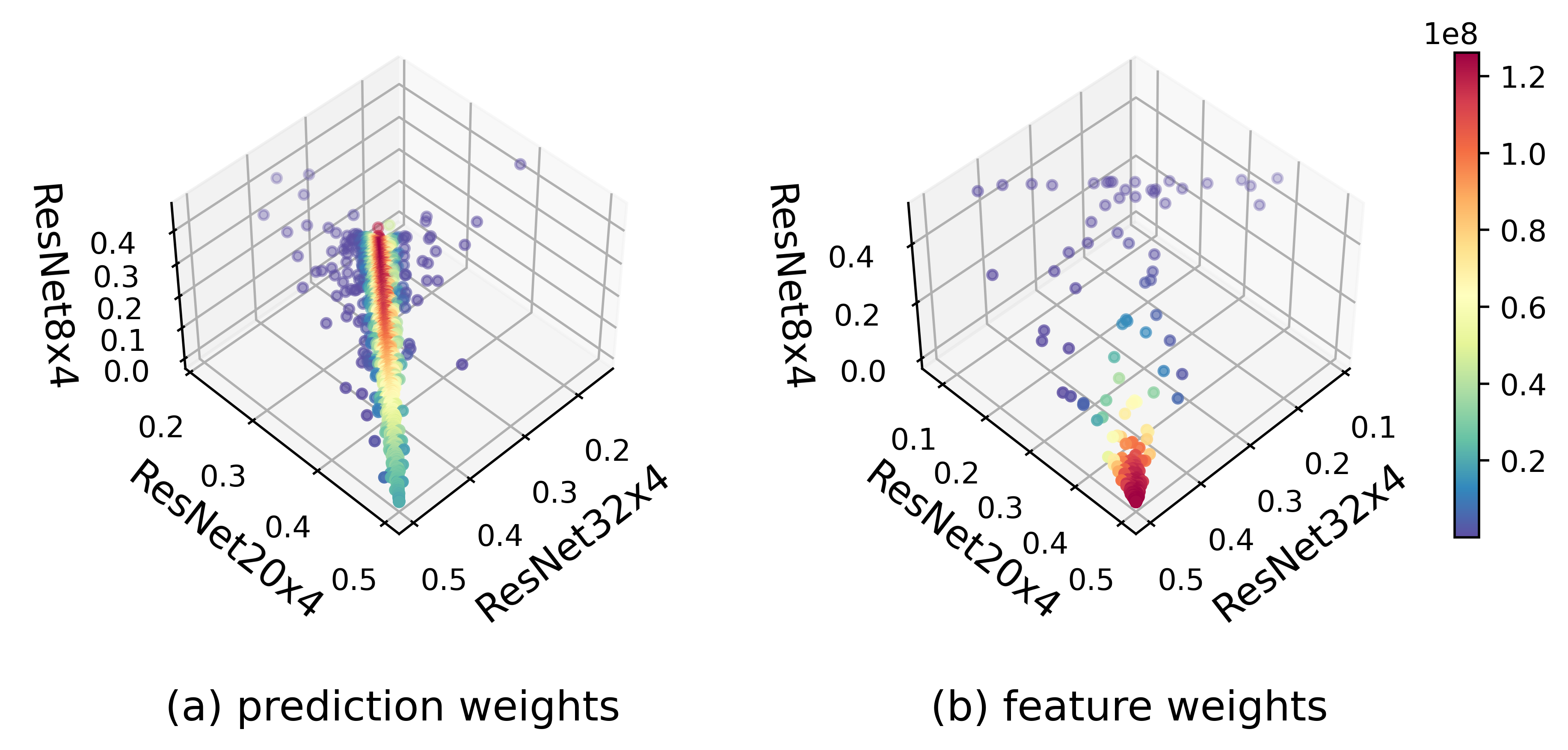}}
    \caption{The visualization results of learned weights by CA-MKD on each training sample.} 
    \label{vis}
    \vspace{-0.3cm}
\end{figure}
\subsection{Results on the Different Teacher Architectures}

Table~\ref{table:MKD_d} shows the results of training a student (VGG8) with three different teacher architectures, i.e., ResNet8x4, ResNet20x4 and ResNet32x4. We find the student accuracy becomes even higher than that of training with three ResNet32x4 teachers, which may be attributed to that the knowledge diversity is enlarged in different architectures. 

Since the performance of ResNet20x4/ResNet32x4 is better than that of ResNet8x4, we could reasonably believe that for most training samples, the student will put larger weights on predictions from the former two rather than the latter one, which is verified in Figure~\ref{vis}. Moreover, our CA-MKD can capture those samples on which the predictions are more confident by ResNet8x4, and assign them dynamic weights to help the student model achieve better performance.

\subsection{Impact of the Teacher Number}

\begin{figure}[t]
    \centering
    \centerline{\includegraphics[width=8.5cm]{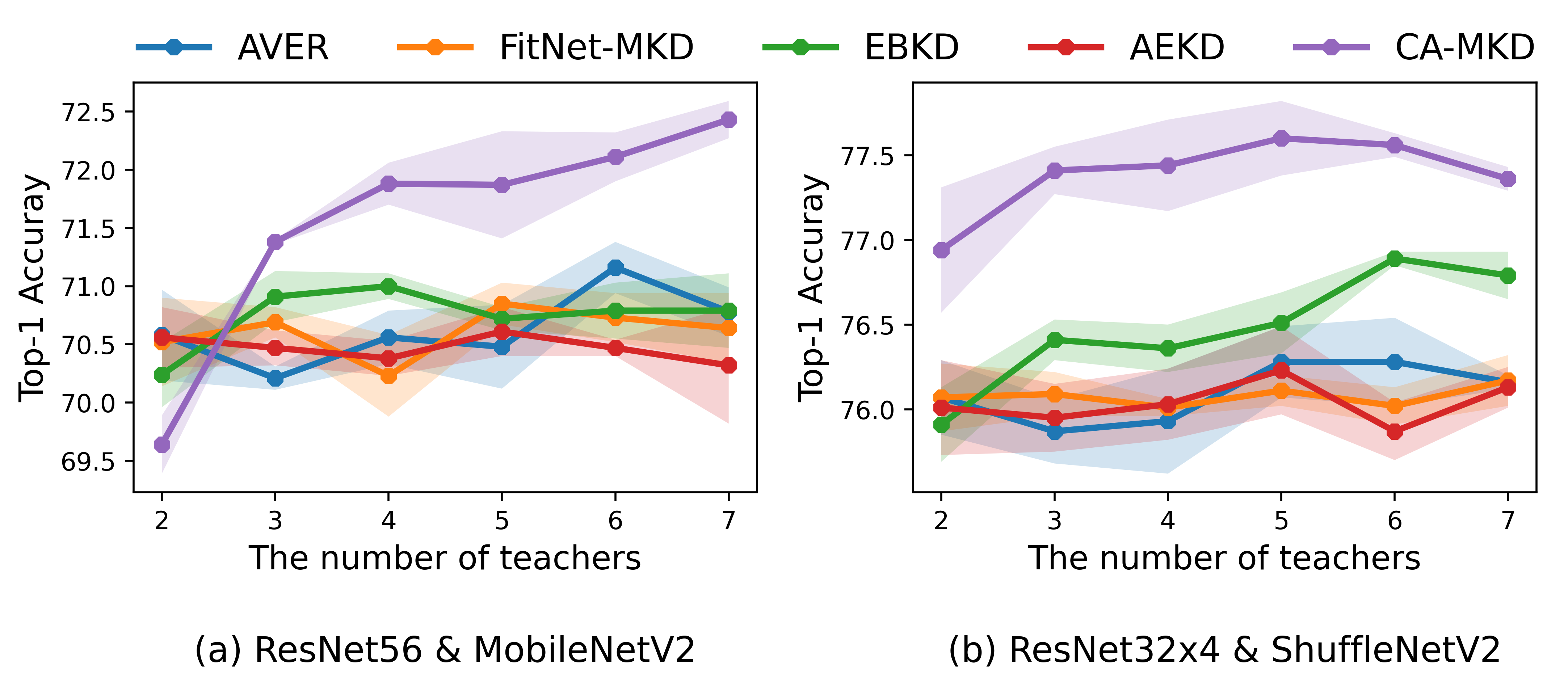}}
    \caption{The effect of different teacher numbers.} 
    \label{Figure 4}
    \vspace{-0.3cm}
\end{figure}

As shown in Figure~\ref{Figure 4}, the student model trained with CA-MKD generally achieves satisfactory results. For example, on the ``ResNet56 \& MobileNetV2" setting, the accuracy of CA-MKD increases continually as the number of teachers increases and it surpasses the competitors with three teachers even those competitors are trained with more teachers. 

\subsection{Ablation Study}

We summarize the observations from Table~\ref{table:ablation} as follows:

(1) avg weight. Simply averaging multiple teachers will cause 1.67\% accuracy drop, which confirms the necessity of treating different teachers based on their specific quality. 

(2) w/o $\mathcal{L}_{inter}$. The accuracy will appear considerably reduction as we remove the Equation~(\ref{eq:inter}), demonstrating the intermediate layer contains useful information for distillation.

(3) w/o $w_{inter}^{k}$. we directly use the $w_{KD}^{k}$ obtained from the last layer to integrate intermediate features. The lower result indicates the benefits of designing a separate way of calculating weights for the intermediate layer.

\begin{table}[t]
\centering
\caption{Ablation study with VGG13 \& MobileNetV2.}
\label{table:ablation}
\resizebox{0.98\columnwidth}{!}{
\begin{tabular}{cccc}
\toprule
avg weight &w/o $\mathcal L_{inter}$ &w/o $w^{k}_{inter}$ &CA-MKD\\
\midrule
67.74$\pm$0.87 &68.11$\pm$0.02 &68.82$\pm$0.63 &\bf69.41$\pm$0.20\\
\bottomrule
\end{tabular}}
\end{table}

\section{CONCLUSION}
\label{sec:page}
In this paper, we introduce confidence-aware mechanism on both predictions and intermediate features for multi-teacher knowledge distillation. The confidence of teachers is calculated based on the closeness between their predictions or features and the ground-truth labels for the reliability identification on each training sample. With the guidance of labels, our technique effectively integrates diverse knowledge from multiple teachers for the student training. Extensive empirical results show that our method outperforms all competitors in various teacher-student architectures.


\bibliographystyle{IEEEbib}
\bibliography{strings,refs}

\newpage
\renewcommand\thesection{\Alph{section}}
\renewcommand\thesection{\arabic{subsection}}
\section*{Appendix}
\subsection{The Detailed Description of $w_{inter}^{k}$}
To stable the knowledge transfer process, we design the student to be more focused on imitating the teacher with a similar feature space and $w^{k}_{inter}$ indeed serves as such a similarity measure representing the discriminability of a teacher classifier in the student feature space. 

\begin{figure}[h]
    \centering
    \centerline{\includegraphics[width=8.5cm]{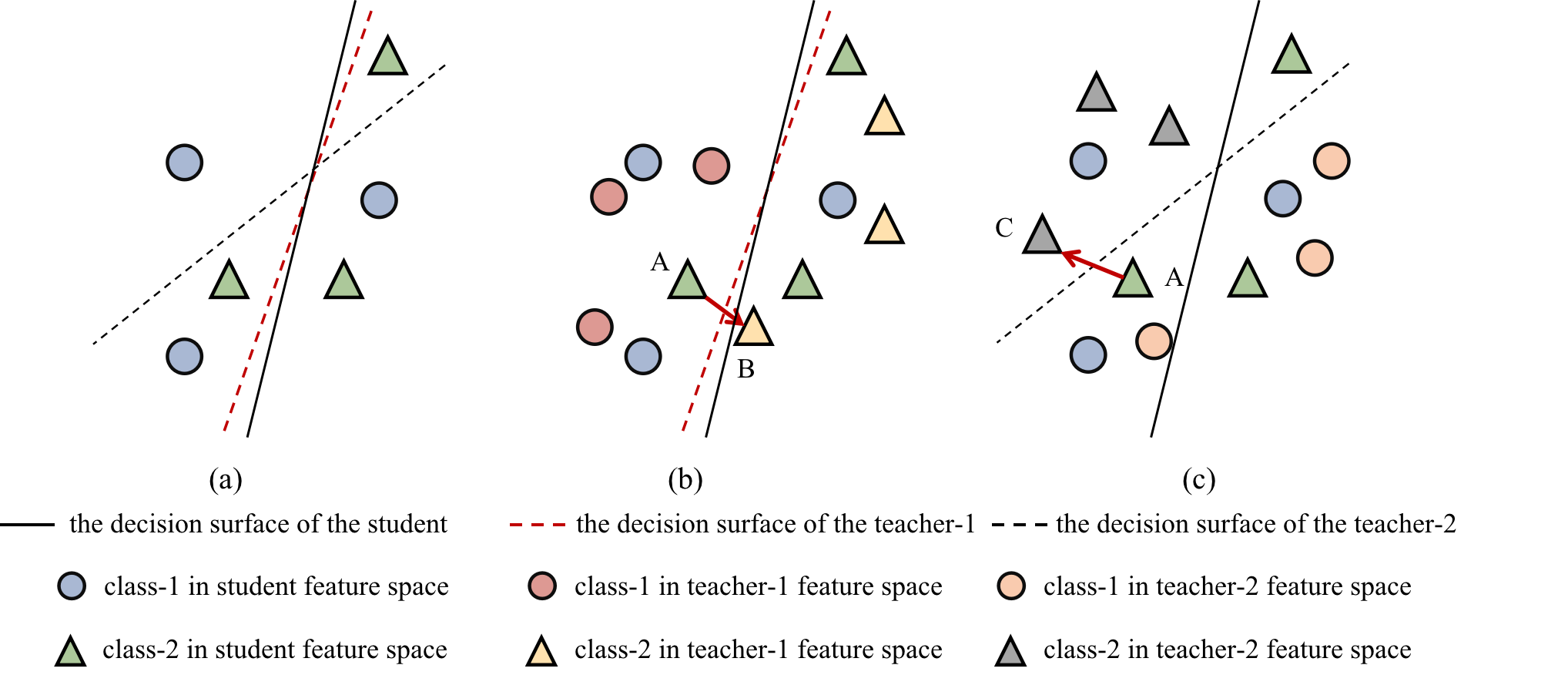}}
    \caption {The comparison of  teacher-1 and teacher-2 classifiers. }
    \label{fig:1}        
\end{figure}

A more detailed discussion is presented in the following paragraphs.As shown in Figure~\ref{fig:1}, samples belonging to class-1 and class-2 are depicted as circles and triangles, respectively. 
Although the decision surfaces of teacher-1 (in Figure~\ref{fig:1}(b)) and teacher-2 (in Figure~\ref{fig:1}(c)) correctly classify these samples in their own feature spaces, their discriminability in the student feature space is different (in Figure~\ref{fig:1}(a)). 

In order to stabilize the whole knowledge transfer process, we expect the student to pay more attention to mimicking the teacher with a similar feature space. In this sense, we conclude that teacher-1 for the student is more suitable since its decision surface performs better compared to that of teacher-2 in the student feature space, as shown in Figure~\ref{fig:1}(a). 

Suppose the point A, B, C are the extracted features of the same sample in the feature space of student, teacher-1 and teacher-2, respectively. 
If we move the student feature (point A) towards the feature from teacher-1 (point B), point A will be correctly classified by the student's own classifier with only minor or even no adjustment. But if we move the student feature (point A) towards the feature from teacher-2 (point C), it will become even harder to be correctly classified by the student, which may disrupt the training of the student classifier and slow down the model convergence.

\subsection{Additional Dataset experiments}
we add more experiments on Dogs and Tinyimagenet datasets to further verify the effectiveness of our proposed CA-MKD. 
Table~\ref{table:mkd} and Table~\ref{table:single} show that our CA-MKD can consistently surpass all the competitors in two more challenging datasets.

The hyper-parameters for the Tinyimagenet dataset are exactly the same as those of CIFAR-100 in our submission. Another dataset (Dogs) contains fine-grained images with larger resolutions, which requires a different training procedure. We follow the setting of a previous work \cite{yun2020regularizing}. \\

  \centering
     \makeatletter\def\@captype{table}\makeatother\caption{Top-1 test accuracy of CA-MKD compared to multiple-teacher knowledge distillation methods.}
     \label{table:mkd}
       \begin{tabular}{cccc} 
\toprule
Dataset &Dogs &Tinyimagenet \\
\midrule
\multirow{2}*{Teacher}&ResNet34 &ResNet32x4 \\
&64.76$\pm$1.06 &53.38$\pm$0.11 \\
\midrule
\multirow{2}*{Student}&ShuffleNetV2x0.5 &VGG8 \\
&59.36$\pm$0.73 &44.40$\pm$0.15\\
\midrule
AVER  &64.49$\pm$0.16 &47.82$\pm$0.15 \\
FitNet-MKD  &64.11$\pm$0.80 &47.82$\pm$0.05 \\
EBKD &64.32$\pm$0.23 &47.20$\pm$0.10 \\
AEKD &64.19$\pm$0.34 &47.62$\pm$0.38 \\
\midrule
CA-MKD &\bf65.19$\pm$0.23 &\bf49.55$\pm$0.12 \\
\bottomrule
	\end{tabular}

\vspace{0.5cm}

   \centering
        \makeatletter\def\@captype{table}\makeatother\caption{Top-1 test accuracy of CA-MKD compared to single-teacher knowledge distillation methods.}
        \label{table:single}
        \begin{tabular}{cccc} 
\toprule
Dataset &Dogs &Tinyimagenet \\
\midrule
\multirow{2}*{Teacher}&ResNet34 &ResNet32x4 \\
&65.97 &53.45 \\
\midrule
\multirow{2}*{Student}&ShuffleNetV2x0.5 &VGG8 \\
&59.36$\pm$0.73 &44.40$\pm$0.15\\
\midrule
KD  &63.90$\pm$0.08 &47.42$\pm$0.07 \\
FitNet  &62.45$\pm$0.61 &47.24$\pm$0.28 \\
AT &63.48$\pm$0.60 &45.73$\pm$0.05 \\
VID &64.45$\pm$0.23 &47.76$\pm$0.08 \\
CRD &64.61$\pm$0.17 &48.11$\pm$0.07 \\
\midrule
CA-MKD &\bf65.19$\pm$0.23 &\bf49.55$\pm$0.12 \\
\bottomrule
\end{tabular}
\end{document}